\documentclass[11pt,twocolumn]{article}
\pdfoutput=1
\usepackage{booktabs} 
\usepackage{graphicx}
\usepackage[numbers]{natbib}
\usepackage{amsmath}
\usepackage{amssymb}
\usepackage{multirow}
\usepackage{hyperref}
\usepackage{breqn}

\allowdisplaybreaks

\usepackage{pifont}
\newcommand{\cmark}{\ding{51}}%

\usepackage{authblk}
\title{Unsupervised Rank-Preserving Hashing for Large-Scale Image Retrieval}

\author[1]{Svebor Karaman}
\author[1]{Xudong Lin}
\author[2]{Xuefeng Hu}
\author[1]{Shih-Fu Chang}

\affil[1]{Columbia University, New York, NY, USA}
\affil[2]{University of Southern California, Los Angeles, CA, USA}
\date{}
\begin{document}

\maketitle

\begin{abstract}
We propose an unsupervised hashing method which aims to produce binary codes that preserve the ranking induced by a real-valued representation.
Such compact hash codes enable the complete elimination of real-valued feature storage and allow for significant reduction of the computation complexity and storage cost of large-scale image retrieval applications. 
Specifically, we learn a neural network-based model, which transforms the input representation into a binary representation.
We formalize the training objective of the network in an intuitive and effective way, considering each training sample as a query and aiming to obtain the same retrieval results using the produced hash codes as those obtained with the original features.
This training formulation directly optimizes the hashing model for the target usage of the hash codes it produces.
We further explore the addition of a decoder trained to obtain an approximated reconstruction of the original features.
At test time, we retrieved the most 
promising database samples with an efficient graph-based search procedure using only our hash codes and perform re-ranking using the reconstructed features, thus without needing to access the original features at all.
Experiments conducted on multiple publicly available large-scale datasets show that our method consistently outperforms all compared state-of-the-art unsupervised hashing methods and that the reconstruction procedure can effectively boost the search accuracy with a minimal constant additional cost. 
\end{abstract}
\section{Introduction}
\label{sec:introduction}

\begin{figure}[tb]
\includegraphics[width=\columnwidth,,bb=0 0 748 690]{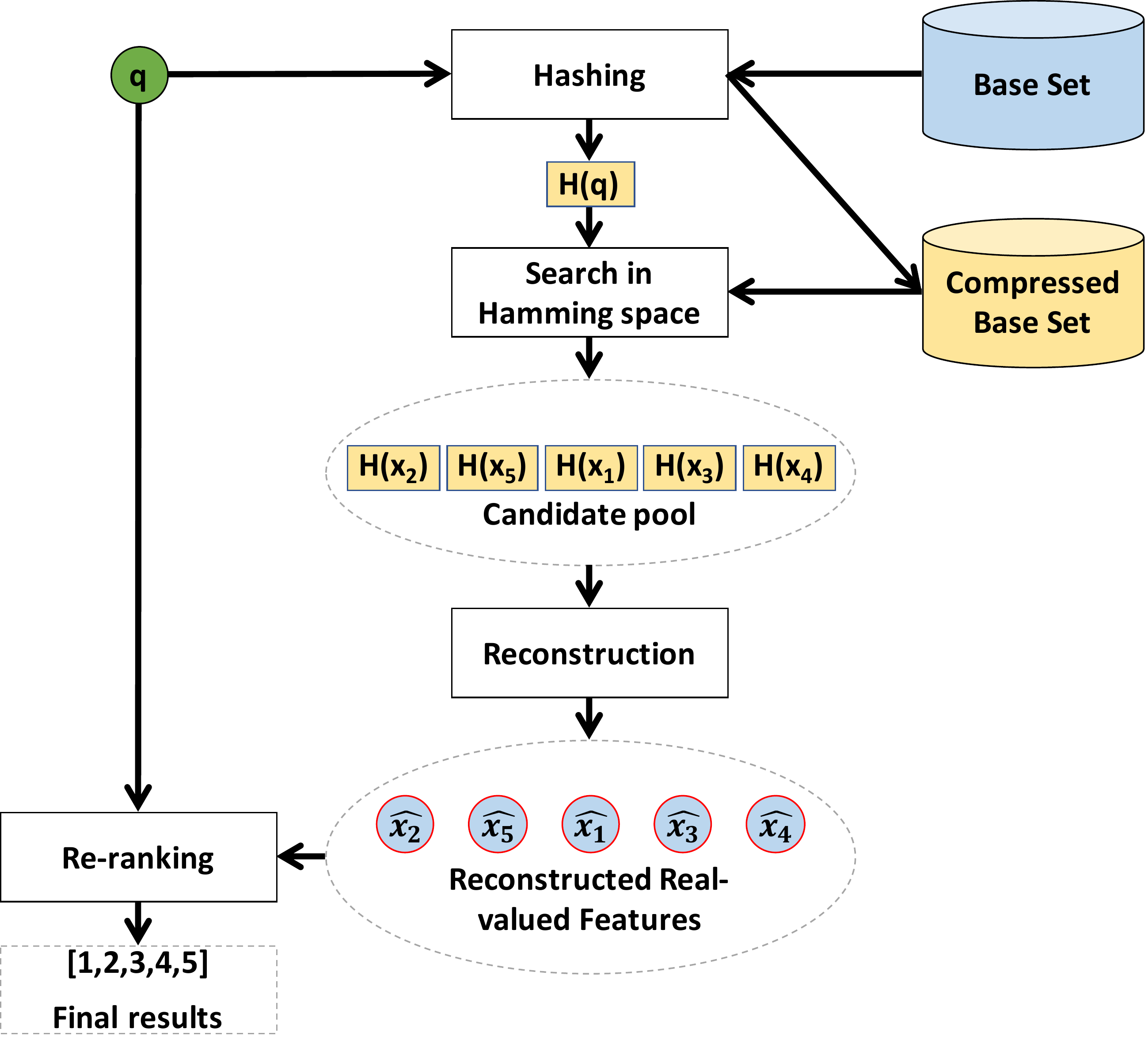}
\caption{An illustration of the search strategy with our URPH model and its decoder. We first compress the database to binary hash codes offline. At test time, given a query $q$, we hash it and obtain a set of candidates by searching in the hamming space. We then use the decoder to reconstruct an approximation of these candidate samples' original features and re-rank them using the real-valued query feature.\label{fig:URPH_asym_search}}
\end{figure}

Researchers in computer vision have developed many methods~\cite{lowe2004distinctive,krizhevsky2012imagenet,li2018patternnet} 
to extract representations that can capture the most important information of an image.
These representations are commonly used in content-based retrieval systems enabling searching for similar images in a large database.
However, for a large database, the cost associated with an exhaustive search and of storing all the uncompressed features both become prohibitive.
Therefore, the development of methods that can efficiently perform a high-accuracy approximate nearest neighbour search (ANN) and of methods to compress the original representation while preserving their discriminability are both active research topics.

The ANN problem has been an active research topic for decades~\cite{bentley1975multidimensional,charikar2002similarity}, but an emerging trend in the last few years is to exploit a graph structure to perform the search. 
The Hierarchical Navigable Small-World (HNSW) method~\cite{malkov2018efficient} is now dominating the ANN benchmarks\footnote{https://github.com/erikbern/ann-benchmarks}. 
The main drawback of such graph-based method is that they require storing the original features.
On the other hand, many works have proposed to learn a hashing model that can compress a feature by producing a binary representation.  
However, most of the literature has focused on settings where the produced hash codes have low dimensions (i.e. 64 bits or less) and thus suffered from limited discriminative power compared to the original features.
In this setting, hashing is often used as a first step in the search process. 
Namely, computing exhaustively all the hamming distances between a query hash code and all database hash codes and gathering a small set (e.g. few thousands) of samples that have the smallest hamming distances.
A second step, often referred to as \emph{re-ranking}, is to get the corresponding original features of the few thousands gathered samples and compute the exact distances with the query feature.
This re-ranking step can be slow if the database samples features are read from disk and would still require a substantial amount of memory to store all the features to allow faster access.
Furthermore, many hashing methods~\cite{liu2012supervised,song2015top,song2015rank} rely on the labels of the training samples to optimize the hash functions. 
This setting requires potentially a large number of labels and its evaluation has been questionable~\cite{sablayrolles2017should} if the same labels are used in the training and test phase. 
There is also recent work~\cite{douze2018link} applying the product quantization coding method to HNSW graph-based indexing. 
Their problem setting is more similar to those hashing methods that produce low-dimension hash codes, where compression ratio is more important than search performance. 
Their product quantization based method doesn't have the advantage of fast hamming distance computation of hash codes. 
Due to these differences, we consider it as a related but different problem. 

In this work, our goal is to totally get rid of the original features and perform the ANN search relying only on the compressed representation.
We assume that the features have been heavily optimized for the target task, and our goal is to preserve their performance when compressing them into a binary representation without being given any labeled data. 
We therefore propose to learn an unsupervised hashing model that produces medium length hash codes (e.g. from 64 to 512 bits), that are optimized directly for the search task by aiming to 
preserve the rank order of retrieval within any set of samples.
We further show that a decoder, that learns to reconstruct an approximation of the original features from hash codes, can be exploited to perform re-ranking without requiring the original features. 
We first collect a candidate pool of the most promising database samples using only the hash codes, then reconstruct an approximation of their features and re-rank the candidates by comparing the query feature to their reconstructed features.
Our search procedure is illustrated in Figure~\ref{fig:URPH_asym_search}.

Our approach is an effective solution to solve the large-scale content-based retrieval problem while allowing to fully discard the original features. 
Experiments conducted on million-scale publicly available datasets show the consistent superiority of our method compared to many state-of-the-art hashing methods (with relative performance gains of up to 35\%).
Furthermore, our method can be integrated with the graph-based search method~\cite{malkov2018efficient} overcoming its main limitation which is its memory usage. For example, for the SIFT1M~\cite{jegou2011product} dataset, the proposed method can reduce the memory requirement to store the samples from 488MB to 34MB only.

\section{Related works}
Many approaches have been proposed to learn to produce binary 
codes~\cite{charikar2002similarity,liu2012supervised,gong2013iterative,song2015top,song2015rank,duan2018graphbit,rao2016diverse,yang2017discrete} in past decades. 
In this section, we are going to review the rank-preserving~\cite{wang2013order,wang2013learning,zhao2015deep,douze2016polysemous}, 
and reconstruction-related hashing methods~\cite{salakhutdinov2009semantic,kulis2009learning,carreira2015hashing,hu2018deep,hu2018hashing,duan2018minimizing,jegou2011searching,jegou2012anti,douze2016polysemous,douze2018link}, which are the most relevant to our proposed method.

\subsection{Rank-preserving Hashing}

Consider a simple retrieval task, which includes
$$D:=\{x_1,...,x_N\}\subset A$$
$$S(q,x):Q \times D \mapsto \mathbb{N}^+$$
where $D$ is a set of items (i.e. the database or the base set) from the feature space $A$, and $S$ is a ranking function 
that gives the rank of items in $D$ 
given a query $q$ from the query set $Q$. 
Then, a rank-preserving hashing method can be defined as a function $H_S$ that map $D$ into the hamming space where the rank $S$ is preserved, namely,
$\forall q\in Q, \forall x\in D, S(q, x) = S(q, H_S(x))$.
In terms of the ranking function $S$, rank-preserving hashing methods can be divided into two categories based on whether $S$ comes from supervision or just the intrinsic properties of the data.

Supervised rank-preserving hashing methods require having semantic labels for the training samples, and use these labels to define the ground-truth ranks of samples. 
Different formulation to penalize falsely ordered samples in hamming space have been proposed.
\cite{wang2013learning} borrows the idea of list-wise supervision from learning to rank, and maximize the product of triplet based similarity scores in the original space and the hamming space. 
\cite{wang2015ranking} proposes to optimize the Normalized Discounted Cumulative Gain (NDCG) measure to preserve the ground-truth relevance obtained from semantic tags.
\cite{zhao2015deep} minimizes the difference of hamming distances in falsely ordered triplets to preserve the rank based on ground-truth NDCG. 
\cite{song2015top} proposes to construct a triplet with a query, a similar sample and a dissimilar sample and preserve this triplet-wise relative distance relationships to learn the hash codes. 

Unsupervised methods, like ours, usually assume the rank in the original space is the rank we want to preserve. 
\cite{wang2013order} proposes to learn hash functions by an order alignment procedure between the original space and the hamming space, which is solved as multiple binary classification problems. 
\cite{douze2016polysemous} also utilizes a ranking loss as part of their objective function, when trying to map centroids to their unsigned integer identifier/binary code, by penalizing distances  in the hamming space that induce a wrong order. 
Although several works have been proposed on ranking-based hashing, to the best of our knowledge, unsupervised rank-preserving hashing method with a neural network hasn't been explored.

\subsection{Reconstruction-related hashing}

\begin{figure*}[tb]
\includegraphics[width=\textwidth,bb= 0 0 2267 566]{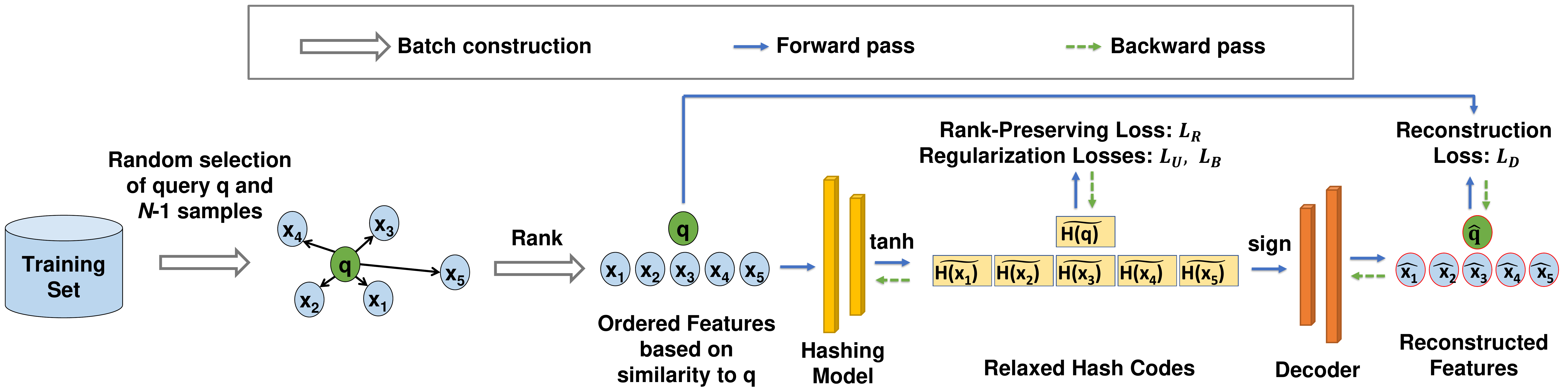}
\caption{Illustration of our URPH training procedure. We first construct training batches by randomly selecting a query $q$ and $N-1$ samples from the training set, where $N$ is the batch size. 
We order the features based on their similarity to $q$ and compute their hash codes.
We utilize the proposed rank-preserving loss and regularization losses to learn a relaxed hashing model. 
Then we use the binarized hash codes to learn a decoder which reconstructs an approximation of the original features, e.g. $\hat{x_1}$, from the hash codes, e.g. $H(x_1)$. 
\label{fig:pipeline_train}}
\end{figure*}

Several works have defined hashing methods that aim to reconstruct an original representation.
Semantic hashing~\cite{salakhutdinov2009semantic} uses a restricted Boltzmann machine with reconstruction as supervision to learn a hashing model. 
In \cite{kulis2009learning}, some assumptions about the data distribution are removed and the reconstruction error is defined as the difference of scaled distance between the original space and the hamming space. 
\cite{carreira2015hashing} proposes to use the method of auxiliary coordinate to optimize an auto-encoder, which first encodes images into binary codes and then decodes binary codes back into images. 
\cite{hu2018deep} uses reconstruction to learn an unified embedding for two modalities. 
\cite{hu2018hashing} also uses similarity as a reconstruction target, and their formulation of angular constraint leverages the cosine similarity to produce more discriminative binary codes. 
\cite{duan2018minimizing} provides a systematic analysis on the relationship of projection and quantization step, and proposes to jointly optimize these two steps by minimizing the reconstruction error. 
All these methods only consider the reconstruction as part of the optimization process to learn the hash model but not at test time.

Reconstructed features can also be used to re-order a small candidate pool retrieved through hamming distance comparison to improve retrieval performance~\cite{jegou2011searching,jegou2012anti,douze2016polysemous,douze2018link} without a large drop of speed.
However, ~\cite{jegou2011searching,douze2016polysemous,douze2018link} either use extra codes or their methods are based on product quantization, which thus cannot exploit low-level instructions (e.g. bit operations) to effectively compute distances as it is possible in the Hamming space. 
\cite{jegou2012anti} utilizes anti-sparse coding to learn an binary representation with the reconstruction error as the optimization goal but encoding each sample is its own separate optimization problem, which makes the encoding step significantly slow. 
They observe that using reconstruction features to re-rank the candidate pool helps improve retrieval performance. 
However, this reconstruction technique may suffer from sub-optimal hash codes as the reconstruction alone does not guarantee that original ranks are well-preserved in the hamming space. 
In our method, we first learn rank-preserving hash codes which are suitable for ranking, and then learn a decoder based on these hash codes.

\section{Our Method}
 Previous hashing methods has achieved significant success on compression and retrieval. However, there are still two problems that have not been fully addressed:
 \begin{itemize}
     \item When we already have a very good representation, how can we preserve as much as possible the ranking induced by these features with binary codes?
     \item Since the hamming distance has limitations when used for ranking both in terms of number of possible distances and distinction between different bits~\cite{douze2016polysemous}, how can we improve the retrieval performance when using binary codes?
 \end{itemize}
 
We propose an unsupervised rank-preserving hashing (URPH) method to better preserve the original features' ranking and a reconstruction scheme for more accurate retrieval. 
An illustration of our training pipeline is shown in Figure~\ref{fig:pipeline_train}. 
In the following parts, we are going to first introduce some preliminary notations and definitions, and then explain the different elements of our method. 

\subsection{Preliminary}
\label{sec:preliminary}

Hashing methods can be effectively used with mainly two search strategies~\cite{wang2018survey}: radius search with hash table lookup or hash code ranking.
As radius search is inefficient for the size of hash codes we use, we are particularly interested in hash code ranking. 
We further aim to exploit an auxiliary data structure to avoid exhaustive search.  
Namely, we will exploit the HNSW~\cite{malkov2018efficient} approach which can significantly reduce the search time while allowing to preserve most of the search performance of any given representation.

Hierarchical Navigable Small World graphs (HNSW)~\cite{malkov2018efficient}, is currently the state-of-art strategy for approximate $k$-nearest neighbour search. 
A Navigable Small World (NSW) graph is a sparse graph (each node having a small number of edges) but where any vertex is reachable from any other vertex in a few hops along the graph edges. 
HNSW combines the power of a NSW graph, and a multi-layer hierarchical structure. 
When creating the HNSW index structure on a database, all the samples are inserted at the bottom layer but have an exponentially decaying probability to be inserted in the top layers.
The ANN search process is a top-down traversal of the hierarchical graph structure.
In the top layers, starting from an entry point, the search process consists in successively moving to the neighbour of the current point closest to the query, if it is closer than the current point.
Once a better neighbour cannot be found in a layer, the current point becomes the entry point of the layer below and the search process is re-iterated in that layer.
Finally, in the bottom layer, a buffer of size \emph{efSearch} is filled by inserting the neighbours of the best unexplored samples of the buffer until all the buffer samples have been explored.
The HNSW search achieves a nearly logarithmic complexity scaling for the approximate $k$-nearest neighbour search, with a very good speed-recall performance.
We refer the reader to~\cite{malkov2018efficient} for additional details.
In this work, we build the HNSW with the binary codes and use its efficient search process instead of the exhaustive search.

Formally, the goal of learning to hash is to learn a mapping $H:\mathbb{R}^n \rightarrow \{0,1\}^m$.  
The binary codes after this mapping are supposed to preserve the ranking in the original space. 
Ideally, the hashing model is supposed to be:
\begin{equation}
    H(x;\theta)=\text{sign} (F(x;\theta))
\end{equation}
where $\text{sign}$ is the sign function and $F(x;\theta):R^n \rightarrow R^m$ is a transformation (e.g. a neural network) and $\theta$ are the parameters of that transformation. 
However, with this hashing model, the loss function will be hard to be optimized by gradient descent due to the sign function not being differentiable. 
To tackle this problem, we have to relax the hashing model in the training time. 
We follow~\cite{wang2018survey} and relax the sign function with the hyperbolic tangent ($\tanh$):
\begin{equation}
    \Tilde{H}(x;\theta)=\tanh (F(x;\theta))
\end{equation}

For simplicity, we will omit the parameters $\theta$ and use $H(x)$ and $\Tilde{H(x)}$ instead of $H(x;\theta)$ and $\Tilde{H}(x;\theta)$ in the rest of the paper. 
Since we relax the hashing model in the training time, we also need to the approximate hamming distance in the training time. 
We use the Euclidean distance as surrogate distance:
\begin{equation}
    d_h(\Tilde{H}(x_i),\Tilde{H}(q)) = ||\Tilde{H}(x_i)-\Tilde{H}(q)||_2
\end{equation}

At test time, we use the trained mapping, without relaxation i.e. the binary codes are fully binary, to compress the original features into hash codes, and search for the nearest neighbours of the query by 
computing hamming distances between hash codes 
leveraging the HNSW index configured\footnote{We set $maxM=32$, $maxM0=64$ and $efSearch=1024$} to have a very good approximation of the exhaustive search performance. 

\subsection{Unsupervised Rank-Preserving Hashing}
Unlike supervised ranking-based hashing methods, there are no semantic labels provided in our setting. 
To obtain ranking in this case, we propose to simulate the search process with the original features to get an ordered candidate pool for one query. 
Therefore our optimization goal is to preserve the ranking of this candidate pool for each query. 
As shown in Figure~\ref{fig:pipeline_train}, during training we construct a mini-batch of batch size $N$ in the following way: (i) randomly select $N$ samples from the training set; (ii) randomly select one sample from them to be the query $q$; (iii) rank the rest of the samples in ascending distance to the query. 

It is obvious that the order is preserved if and only if, for any triplet composed of the query $q$, and two candidates $i$ and $j$, $i$ and $j$ remain in the same order in the hamming space. 
Therefore, when we have a query and $N-1$ candidates ordered with ascending distance to the query in the original space, we want to penalize any triplets in the hamming space that are not in the original order.
Formally, our ranking loss is defined as 
\begin{multline}
L_R = \sum _{i=1}^{N-2} \sum _{j=i+1}^{N-1} w_i  [  d_h(\Tilde{H}(x_i),\Tilde{H}(q)) -\\ d_h(\Tilde{H}(x_j),\Tilde{H}(q))  ]_{+},
\end{multline}
where $q$ is the original feature of the query, and $x_i$ is the feature of the $i^{th}$ candidate, and $[\cdot]_+=\max (\cdot,0)$. 
$w_i$ is a weight to ensure that the model penalizes more the disorder happening at smaller ranks and it is defined as
\begin{equation}
    w_i = \exp{(-\frac{i-1}{N-1})}. 
\end{equation}

We also adopt two widely-used regularization terms~\cite{wang2018survey} in learning to hash methods: bit uncorrelation loss and binarization loss. The bit uncorrelation loss $L_U$ is aimed to make every bit useful by enforcing every pair of bit uncorrelated, and is computed as
\begin{equation}
    L_U = ||\Bar{B}^T\Bar{B}-I||_2,
\end{equation}
$N$ is the number of bits, $\Bar{B}=[\Bar{H}(q);\Bar{H}(x_1);\Bar{H}(x_2);...;\Bar{H}(x_{N-1})]$ is a $N\times M$ matrix consisting of $N$ $M$-bit l2-normalized relaxed hash codes, 
and $I$ is the $M \times M$ identity matrix. 
Note that here we apply l2-normalization to relaxed hash codes before calculating the uncorrelation loss. 
Since hash codes are relaxed, wihtout l2-normalization, there is a trivial sub-optimal case which also produces a small loss value when every bit of the relaxed hash codes is close to zero. 

The binarization loss $L_B$ helps to enforce the model to produce binarized vectors:
\begin{equation}
    L_B = 1 - \frac{1}{NM}\sum _{i=1}^{N}\sum _{j=1}^{M}\Tilde{B}_{ij}^2
\end{equation}
where $\Tilde{B}_{ij}$ denotes the element in $i^{th}$ row and $j^{th}$ column of matrix $\Tilde{B}=[\Tilde{H}(q);\Tilde{H}(x_1);\Tilde{H}(x_2);...;\Tilde{H}(x_{N-1})]$. 
Different from uncorrelation loss, we use relaxed hash codes without l2-normalization for binarization loss. 
As we relax the hashing model and use hyperbolic tangent as an approximation of sign function during training time, this term forces the model to produce hash codes that have values closer to the extremes $\{-1, 1\}$. 

Therefore our final objective for hashing model learning is:
\begin{equation}
    L = \lambda_1*L_R+\lambda_2*L_U+\lambda_3*L_B
\end{equation}
With this objective, we are able to learn a hashing model which produces rank-preserving and uncorrelated hash codes.

\subsection{Reconstructing features from binary codes}
\label{sec:asymmetric_search}
Due to a small capacity of possibilities of distances~\cite{douze2016polysemous}, there will usually be ambiguity or ties~\cite{he2018hashing} in the candidate pool retrieved by hamming distance comparison. 
For example, there may be several nearest neighbours with the same distance in the hamming space, although only one of them is the ground-truth one. 
There may be even worse case that none of the top nearest neighbours in the hamming space is the ground-truth one due to the information loss of hashing. 
To handle this issue, we propose to reconstruct approximation of the original features and compute an asymmetric distance (between the real-valued query feature and the candidates' reconstructed features) for re-ranking the candidate pool.

Formally, from hash codes of the training set, we learn a decoder $D(b;\phi):\{0,1\}^m \rightarrow \mathbb{R}^n$, where $\phi$ are the parameters of the model. 
Let us denote $\hat{x_i} = D(b_i;\phi)$ the reconstructed feature of $x_i$, with $b_i =  H(x_i)$.
We use a l2-norm loss as the optimization objective for the reconstruction:
\begin{equation}
    L_D = \frac{1}{N} \sum _{i=1}^{N} ||x_i-\hat{x_i}||_2
\end{equation}
where $||\cdot||_2$ denotes the l2-norm. Note that here $b_i$ is a binarized code without relaxation, and there is no gradient back-propagated from the decoder to the hashing model.

With a trained hashing model and decoder, our full searching process is shown in Figure~\ref{fig:URPH_asym_search}. 
First, compress the base set with our hashing model. Second, use the hash codes of the query to perform
an approximated search with HNSW and obtain a candidate pool of $K=100$ nearest neighbours. 
Third, reconstruct real-valued features for all the candidates and use these features to compute the asymmetric distance for re-ranking. 
Formally, the asymmetric distance $d_a$ we use to re-rank the candidate pool is defined as:
\begin{equation}
    d_a(q,b_i) = d(q,\hat{x_i})
\end{equation}
where $q$ is the original feature of the query, $b_i$ is the binary code of sample $x_i$ and $\hat{x_i}$ its reconstructed feature, and $d(\cdot,\cdot)$ is the Euclidean distance in our experiments.

\subsection{Discussions}
There are previous works on unsupervised rank-preserving methods~\cite{wang2013order,douze2016polysemous}, and using reconstructed features for re-ranking~\cite{jegou2012anti}. Compared to these, our objectives and optimization strategies are different. 
The usage of neural network and stochastic gradient descent makes it easier for us to optimize a hashing model on a large training set and produce an output that has a relatively large number of bits. 
The fundamental difference lies in the problem formulation, for which we specifically emphasize the ranking ability with compressed representations.

\section{Experiments}
In this section, to stress the effectiveness of our URPH, we evaluate our method on two public large-scale datasets: SIFT1M~\cite{jegou2011product} and Deep1M~\cite{babenko2016efficient}. 
We compare our method with state-of-the-art methods under different settings. 
We also provide a detailed study of our method performance with different network architectures and in the asymmetric search setting.

\subsection{Datasets}
Hashing models are evaluated on various datasets. Supervised hashing methods can be evaluated on image datasets with semantic labels. However, in our setting, we are more interested in preserving the ranking in original space without relying on image labels. 
We assume that we have already good representations from the image content.
Therefore we choose two public, widely-used datasets which have different type of features to evaluate our method.
\begin{itemize}
    \item SIFT1M~\cite{jegou2011product} consists of a training set, a query set and a base set, which are not overlapped. There are respectively $100,000$, $10,000$, and $1,000,000$ samples in these sub-sets. Each sample is a 128-dimension SIFT~\cite{lowe2004distinctive} feature, which is the one of the best hand-crafted local features. We use l2-normalization as a pre-processing step for this dataset.
    \item Deep1M~\cite{babenko2016efficient} consists of a training set, a query set and a base set, which are not overlapped. There are respectively $1,000,000$, $10,000$, and $1,000,000$ samples in these sub-sets. Each sample is a 96-dimension normalized deep feature produced by a deep neural network. Note that Deep1M is a sub-set of the originally proposed Deep1B~\cite{babenko2016efficient}. We sample Deep1M from Deep1B in the same way as described in \cite{douze2018link}.
\end{itemize}

\subsection{Compared Methods and Evaluation Metrics}
We compare our methods with widely-used unsupervised hashing methods:
ITQ~\cite{gong2013iterative}, 
SH~\cite{weiss2009spectral},
IsoH~\cite{kong2012isotropic},
BRE~\cite{kulis2009learning},
KLSH~\cite{kulis2009kernelized},
LSH~\cite{charikar2002similarity},
SpH~\cite{heo2012spherical} and 
USPLH~\cite{wang2010sequential}. 
We obtain their implementations from~\cite{cai2016revisit}, and run all the experiments in our setting. 
Note that ITQ, SH and IsoH all rely on an eigenvalue decomposition of the features' covariance matrix and thus cannot produce hash codes with a number of bits higher than the original features' dimension.
We also compare our method with another rank-preserving method: OPH~\cite{wang2013order}. 
Since we cannot obtain its implementation, we report the comparable results directly from their paper. 
In the following part, we will denote our hashing model as URPH, and our hashing model with reconstruction and asymmetric searching as URPH-RE. We use $k$HL to denote the different architectures we used for our model. Note as mentioned in Section~\ref{sec:introduction}, the compression method using product quantization~\cite{douze2018link} consider a different problem setting and thus are not considered as suitable baselines.
The Anti-Sparse Coding method~\cite{jegou2012anti} has an excessive encoding time of nearly $0.2$s when using $512$ bits, rendering it not applicable for the high speed search setting we are aiming for.

When it is not specified, the default setting of retrieval experiments is the graph-based indexing and searching, namely, using HNSW with $64$ for the max number of node's edges in the bottom layer and $32$ for top layers. 
The buffer size \emph{efSearch}  for the bottom layer search process is set to $1,024$. 
We first use our hashing model to compress the original features into binary codes. 
Afterwards, no real-valued features of the base set will be needed. 
Then the HNSW graph is built using the binary codes. Therefore only the binary codes of the nodes in HNSW graph and each node's edges are stored. 
Finally, we perform searching on the HNSW graph and obtain a candidate pool. 
The returned candidate pool has one hundred samples in our experiments.
Re-ranking is performed, only for the experiments reported in Section~\ref{sec:asymmetric_search_exp},  within that pool of one hundred samples.

We use the widely-used m-Recall-at-K as the metric for our retrieval experiments. 
Formally, m-Recall-at-K over the query set $Q$ is defined as follows:
\begin{equation}
    m\text{-}Recall@K = \frac{1}{|Q|}\sum_{i=1}^{|Q|} \frac{\text{\# of positive samples}}{m}
\end{equation}
where positive samples are the samples that belongs to the true $m$ nearest samples and are in the top $K$ retrieved samples.

\begin{table}[tb]
\resizebox{\columnwidth}{!}{%
    \begin{tabular}{|c|c|c|c|}
    \hline
          & Layer & Configurations & Output Size \\
        \hline
         \multirow{5}{*}{Hashing model} & Input & - & n \\
        & For 1HL and 2HL & fc + elu +BatchNorm & 8*n  \\
        & For 2HL & fc + elu +BatchNorm & 8*m  \\
        & Output & fc & m\\
        & Hashing & Training: tanh; Testing: sign & m\\
        \hline
        \multirow{4}{*}{Decoder} & Input & - & m \\
        & For 1HL and 2HL & fc + elu +BatchNorm & 8*m  \\
        & For 2HL & fc + elu +BatchNorm & 8*n  \\
        & Output & fc +tanh & n\\
        \hline
    \end{tabular}%
}
\caption{The network architecture of our model. "For 1HL and 2HL" means this layer exists when the model has one or two hidden layers. "fc" denotes the fully-connected layer, and "BatchNorm" denotes the batch normalization layer.}
\label{tab:architecture}
\end{table}

\subsection{Implementation and experimental details}

Many previous deep hashing methods only have one or two layers as hashing layers~\cite{zhao2015deep,duan2018graphbit}. 
Therefore, considering efficiency, we use shallow neural networks as our hashing model and decoder.
When compared to other methods in section~\ref{sec:comp_baselines}, our hashing model has one hidden layer, since some of the baselines we compare to are non-linear. 
We also explore the effect of using no or one more hidden layer and of the use of the decoder in Section~\ref{sec:asymmetric_search_exp}. A hidden layer in our model has
an exponential linear unit (elu)~\cite{clevert2015fast}
as an activation function and is always followed by a batch normalization layer~\cite{ioffe2015batch}, which stabilizes the training process. 
The decoder always has a mirrored architecture with the hashing model. 
Details of the network architectures are provided in Table~\ref{tab:architecture}.

To optimize the loss functions of our hashing model and decoder, we utilize the stochastic gradient descent (SGD) method and implement all our method with Tensorflow~\cite{abadi2016tensorflow}. 
We set 
the initial learning rate to $0.001$, the learning rate decay rate to $0.97$ and the learning rate decay frequency to $5,000$ iterations. 
To ensure that the rank-preserving loss dominates the optimization process, we empirically set $\lambda_1$ to the inverse of the first batch ranking loss value, while the $\lambda_2$, and $\lambda_3$ are set $0.5$ and $0.3$ respectively. 
The mini-batch size $N$ is set to $512$.
We observed that the hashing model converges quickly, so we train the hashing model and the decoder simultaneously for $50,000$ iterations. 
We then train only the decoder for another $50,000$ iterations, in this way the generated hash codes are stable and this enables the decoder to properly converge.

\subsection{Comparison with hashing baselines} 
\label{sec:comp_baselines}

In this section, we compare the ANN search performance of our method with unsupervised hashing baselines. 
Note that we here report our method results using \emph{only} the hash codes and \emph{not} the reconstruction process which will be studied in the next section.

\begin{figure*}[htb]
\includegraphics[width=0.297\textwidth,bb=0 0 770 743]{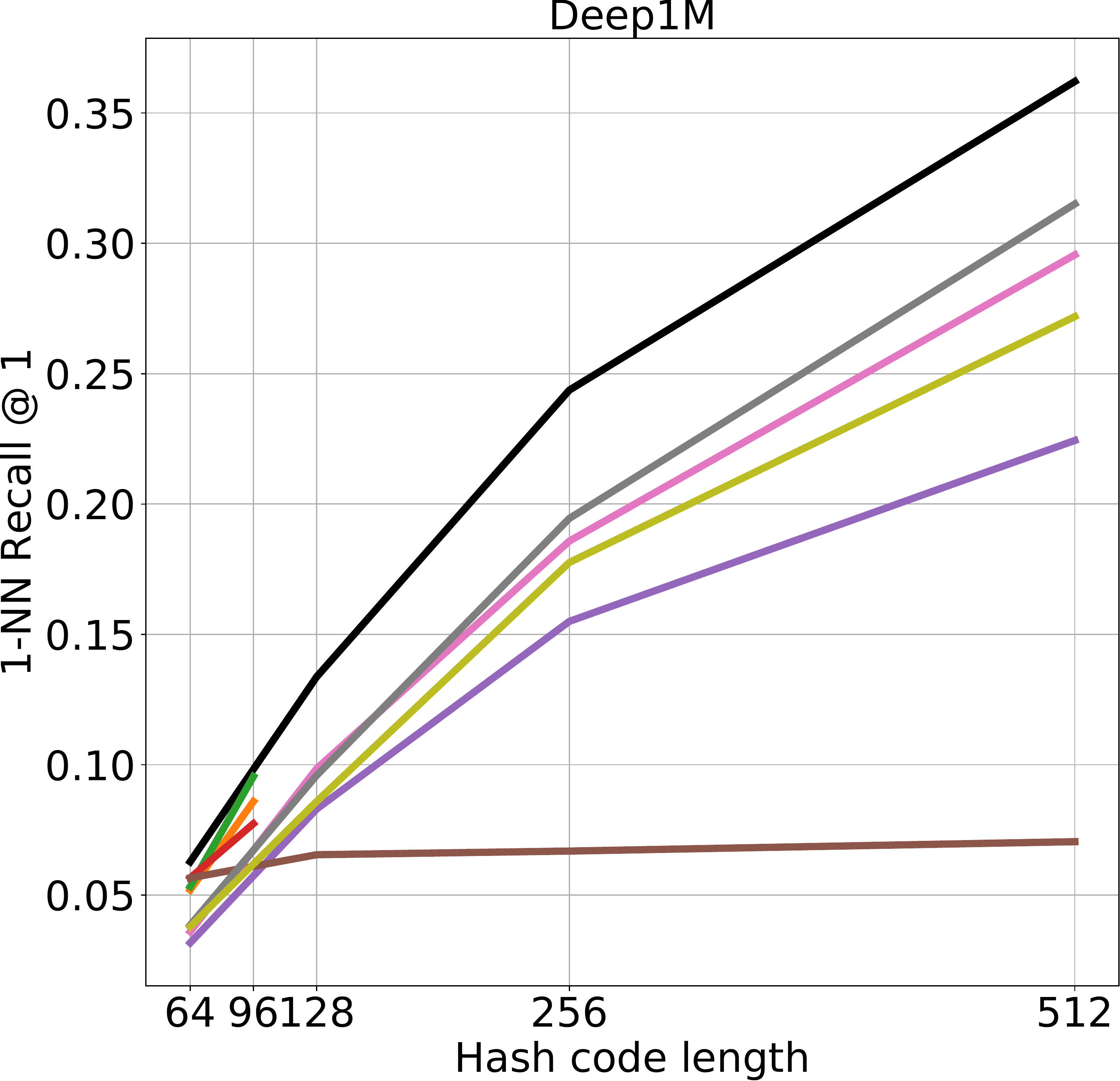}
\includegraphics[width=0.29\textwidth,bb=0 0 2508 2479]{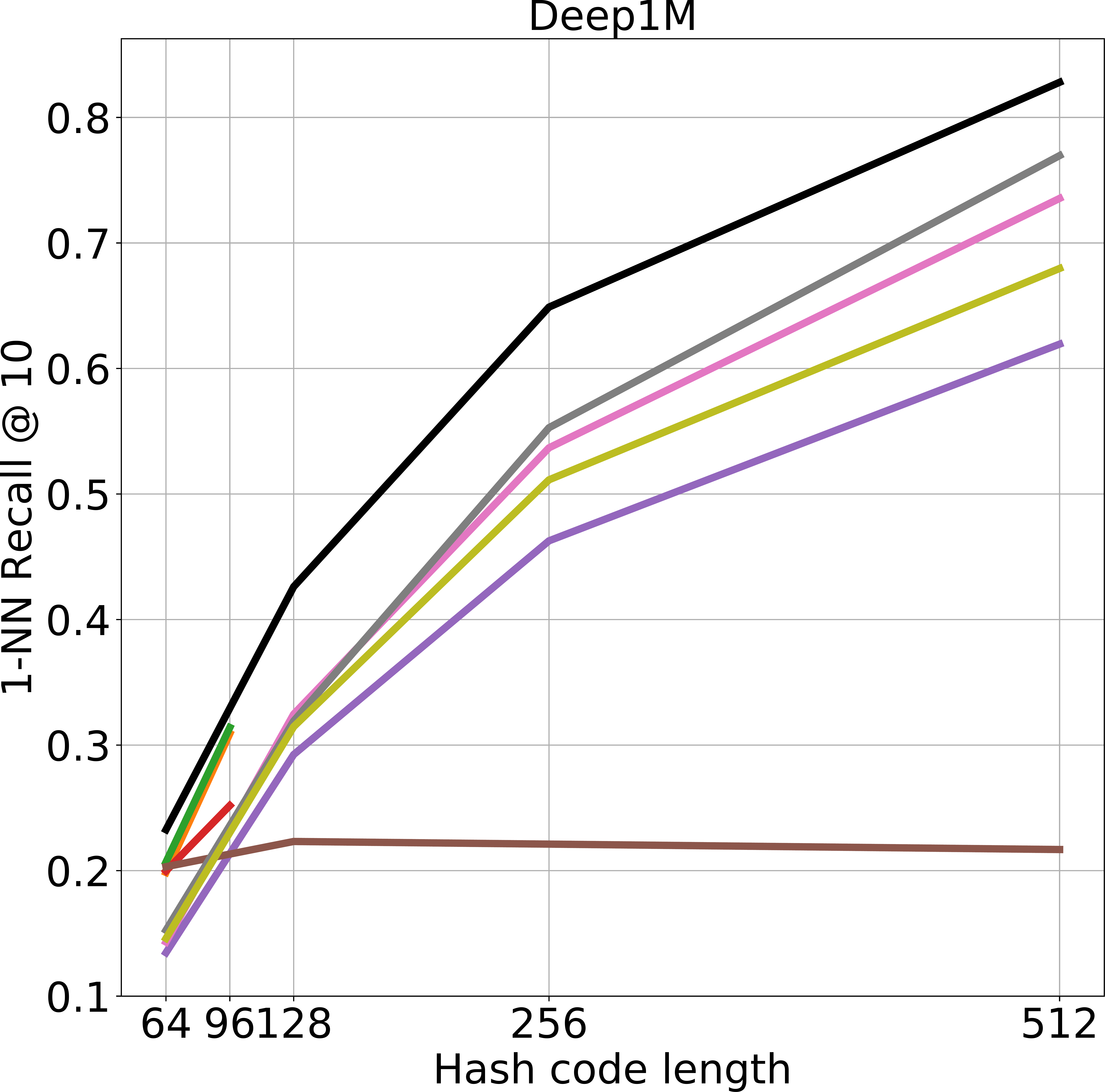}
\includegraphics[width=0.392\textwidth,bb=0 0 1023 759]{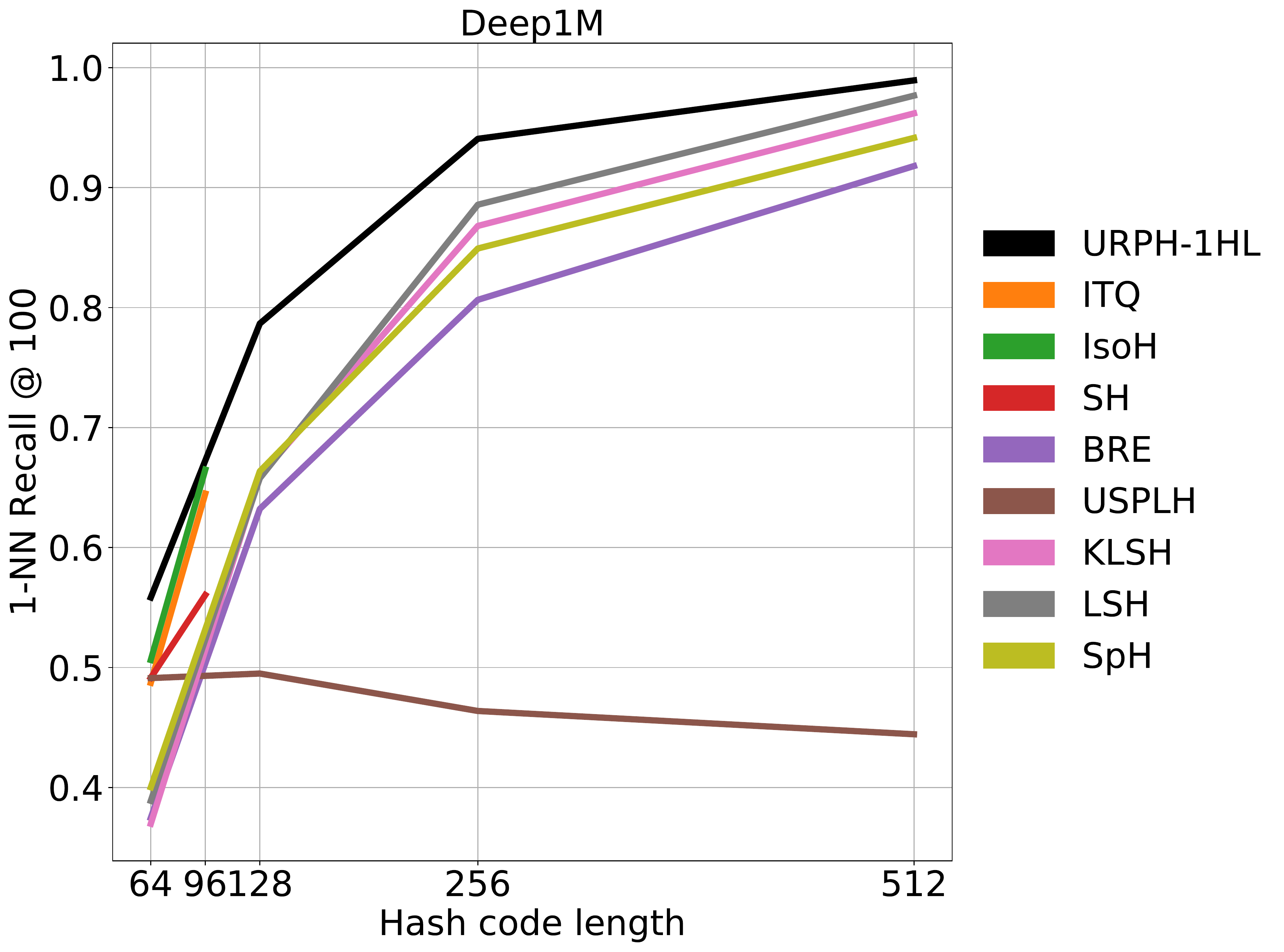}
\caption{1-Recall@1, 10 and 100 on Deep1M with varying hash code lengths.\label{fig:res_recat1_deep1m}}
\end{figure*}

\paragraph{Deep1M:} We report results on the Deep1M dataset in Figure~\ref{fig:res_recat1_deep1m}. Our URPH significantly outperforms previous hashing methods, especially when the number of bits is sufficient. 
For example, when there are 128 bits, the proposed URPH outperforms the best performance among baseline methods by relative gains of  $35.7\%, 31.3\%$, and $19.3\%$ for 1-Recall@(1,10,100); when there are 256 bits, the proposed URPH outperforms the best performance among baseline methods by relative gains of $25.3\%, 17.4\%$, and $6.2\%$ for 1-Recall@(1,10,100).

\begin{figure*}[htb]
\includegraphics[width=0.297\textwidth,bb=0 0 770 744]{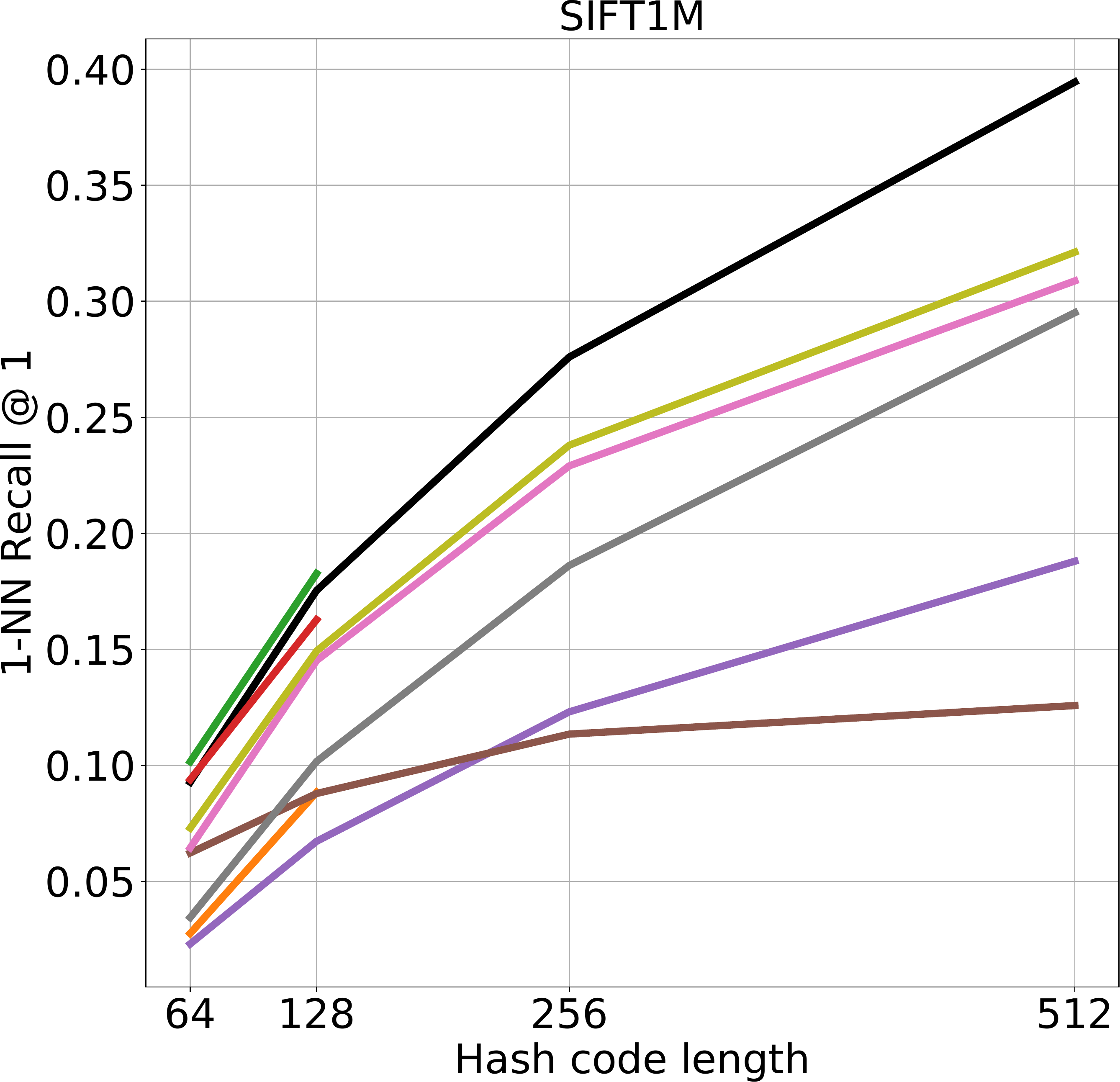}
\includegraphics[width=0.29\textwidth,bb=0 0 752 744]{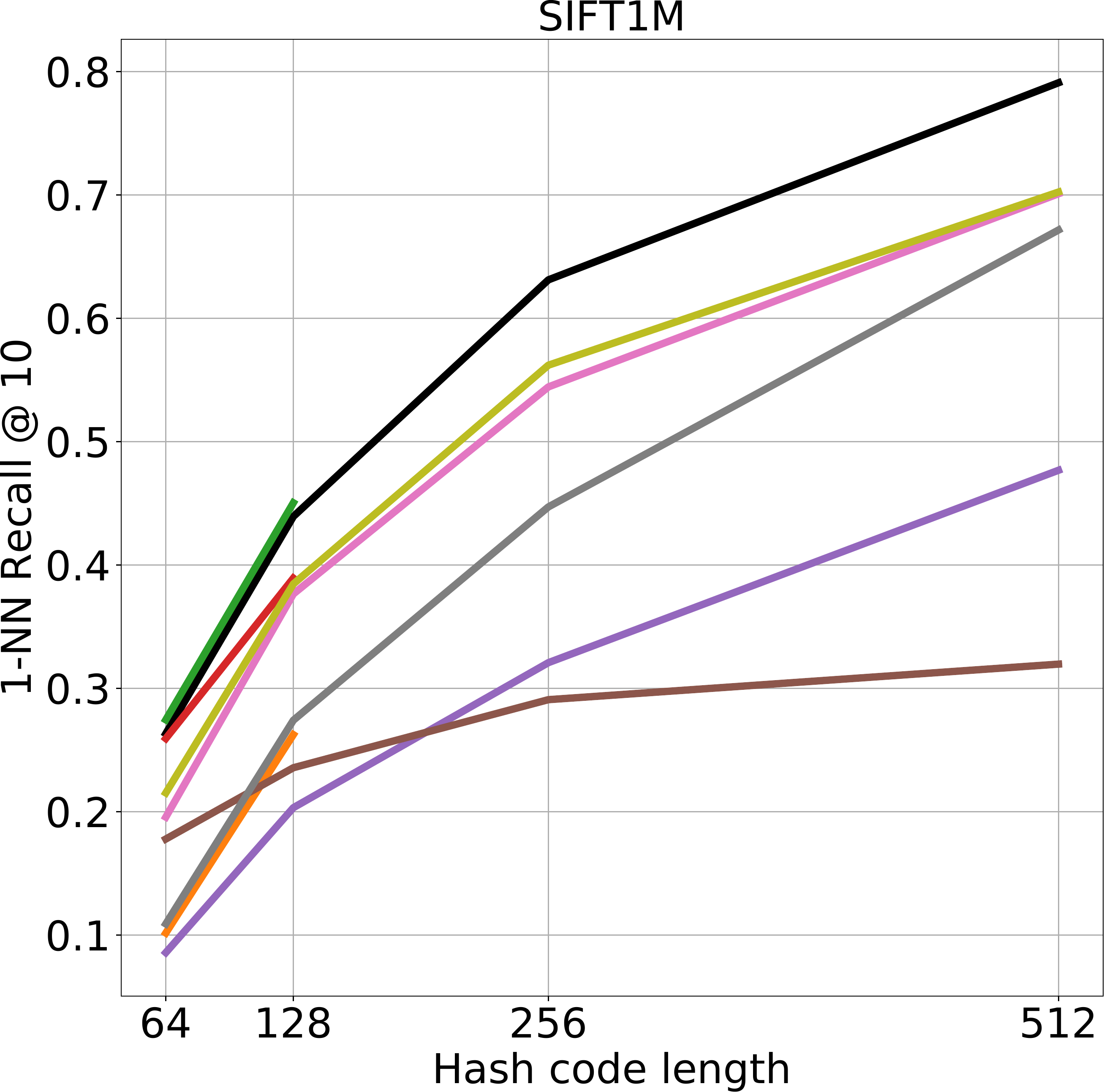}
\includegraphics[width=0.392\textwidth,bb=0 0 1023 759]{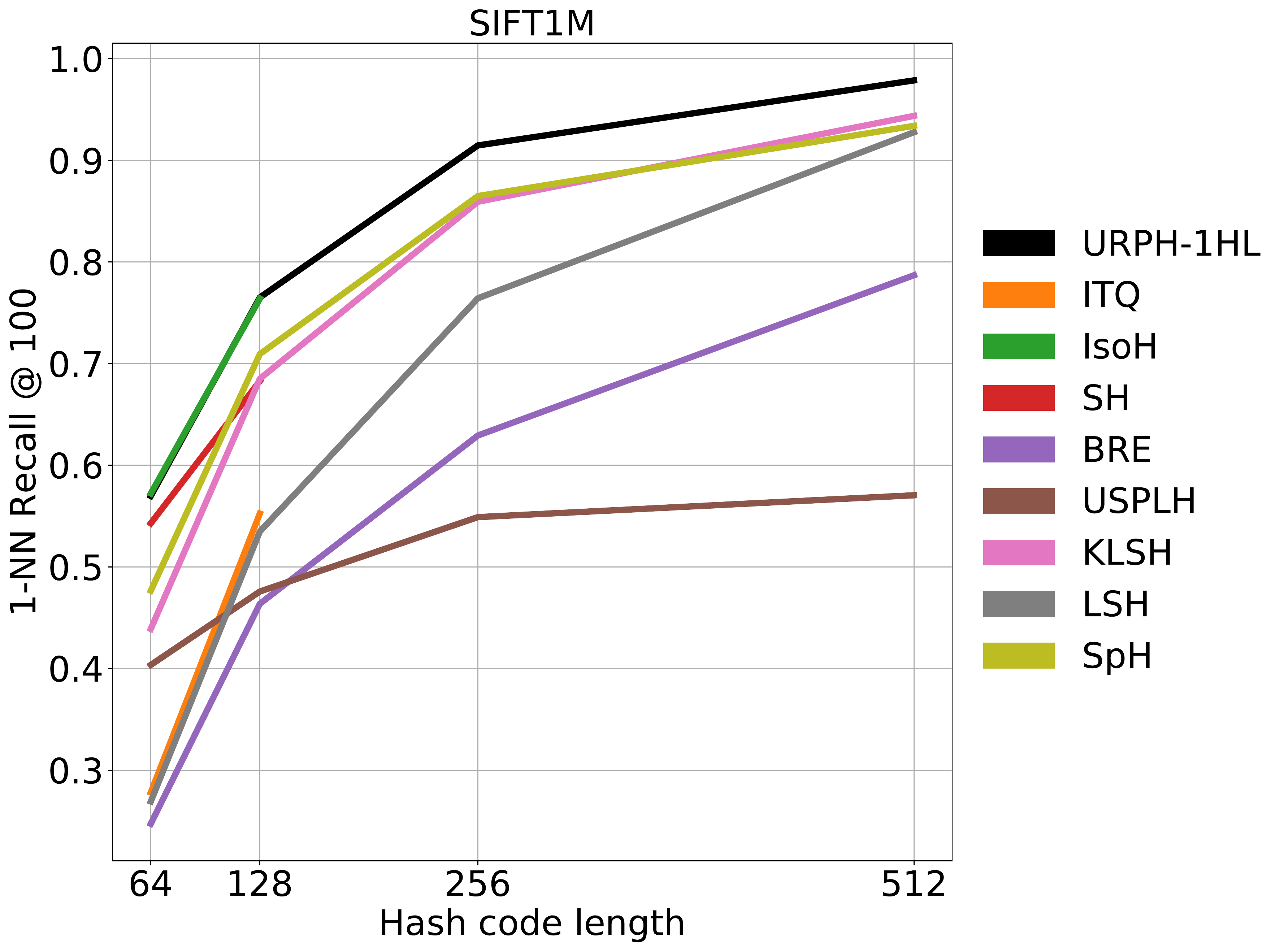}
\caption{1-Recall@1, 10 and 100 on SIFT1M with varying hash code lengths.\label{fig:res_recat1_sift1m}}
\end{figure*}

\paragraph{SIFT1M:}  We report results on the SIFT1M dataset in Figure~\ref{fig:res_recat1_sift1m}. 
Our URPH again outperforms all the baselines by a large margin when the number of bits are sufficient (256, 512), and it can still achieve competitve results when the number of bits is small. 
For example, when there are 256 bits, the proposed URPH outperforms the best performance among baseline methods by relative gains of  $16.0\%,12.3\%$, and $5.76\%$ for 1-Recall@(1, 10, 100); when there are 512 bits, the proposed URPH outperforms the best performance among baseline methods by relative gains of $22.8\%,12.6\%$, and $3.7\%$ for 1-Recall@(1, 10, 100).

We did not find a publicly available implementation of OPH~\cite{wang2013order}, however as they have reported their results on SIFT1M, we obtain their performance numbers by reading the curves in their paper. 
Note that their results are based on exhaustive searching. 
As reported in Table~\ref{tab:comparison_order}, our URPH method consistently outperforms OPH as well.

\begin{table}[tb]
    \centering
    \resizebox{0.9\columnwidth}{!}{\begin{tabular}{c|c|c|c||c|c|c|}\cline{2-7}
          & \multicolumn{3}{c||}{1-Recall@10} & \multicolumn{3}{c|}{1-Recall@100} \\
        \hline
         \multicolumn{1}{|c|}{\# of bits}  & 64 & 128 & 256  & 64 & 128 & 256 \\
        \hline
        \multicolumn{1}{|c|}{OPH}  & 0.20 & 0.31 & 0.37  & 0.51 & 0.68 & 0.81  \\
        \hline
        \multicolumn{1}{|c|}{URPH}  & \textbf{0.26} & \textbf{0.44} & \textbf{0.63}  & \textbf{0.57} & \textbf{0.77} & \textbf{0.91} \\
        \hline
    \end{tabular}}
    \caption{1-Recall@10, 100 on SIFT1M with the varying number of hash bits in [64, 128, 256].}
    \label{tab:comparison_order}
\end{table}

\subsection{Asymmetric search and network structure}
\label{sec:asymmetric_search_exp}

\begin{table}[tb]
    \centering
    \resizebox{\columnwidth}{!}{\begin{tabular}{cc|c|c|c|c||c|c|c|c|}\cline{3-10}
        &  & \multicolumn{4}{c||}{Deep1M} & \multicolumn{4}{c|}{SIFT1M} \\
        \hline
        \multicolumn{1}{|c|}{HL} &  \multicolumn{1}{|c|}{RE} & 64 & 128 & 256 & 512 & 64 & 128 & 256 & 512\\
        \hline
        \multicolumn{1}{|c|}{0} & & 0.055 & 0.13 & 0.239 & 0.366 & 0.083 & 0.169 & 0.272  & 0.387\\
        \hline
        \multicolumn{1}{|c|}{0} & \cmark & 0.088 & \textbf{0.214} & 0.348 & \textbf{0.494} & 0.084 & 0.215 & \textbf{0.346} & \textbf{0.479}\\
        \hline
        \multicolumn{1}{|c|}{1} & & 0.063 & 0.134 & 0.244 & 0.362 & 0.093 & 0.175 & 0.276 & 0.395\\
        \hline
        \multicolumn{1}{|c|}{1} & \cmark & \textbf{0.103} & \textbf{0.214} & \textbf{0.356} & 0.49 & \textbf{0.119} & \textbf{0.226} & 0.342 & 0.451\\
        \hline
        \multicolumn{1}{|c|}{2} & & 0.055 & 0.135 & 0.24 & 0.363 & 0.076 & 0.165 & 0.274 & 0.385\\
        \hline
        \multicolumn{1}{|c|}{2} & \cmark & 0.083 & 0.181 & 0.317 & 0.45 & 0.1 & 0.21 & 0.325 & 0.394\\
        \hline
        
    \end{tabular}}
    \caption{1-Recall@1 with different number of hidden layers (HL) and with 
    re-ranking with reconstructed features (RE) or not, while varying the number of bits in [64, 128, 256, 512].}
    \label{tab:res_asym_net_at1}
\end{table}

\begin{table}[tb]
    \centering
    \resizebox{\columnwidth}{!}{\begin{tabular}{cc|c|c|c|c||c|c|c|c|}\cline{3-10}
        &  & \multicolumn{4}{c||}{Deep1M} & \multicolumn{4}{c|}{SIFT1M} \\
        \hline
        \multicolumn{1}{|c|}{HL} &  \multicolumn{1}{|c|}{RE} & 64 & 128 & 256 & 512 & 64 & 128 & 256 & 512\\
        \hline
        \multicolumn{1}{|c|}{0} & & 0.201 & 0.41 & 0.641 & 0.822  & 0.226 & 0.431 & 0.628 & 0.795\\
        \hline
        \multicolumn{1}{|c|}{0} & \cmark & 0.309 & 0.593 & 0.819 & \textbf{0.94} & 0.276 & 0.552 & 0.753 & \textbf{0.891}\\
        \hline
        \multicolumn{1}{|c|}{1} & & 0.233 & 0.426 & 0.649 & 0.828 & 0.263 & 0.439 & 0.631 & 0.791\\
        \hline
        \multicolumn{1}{|c|}{1} & \cmark & \textbf{0.354} & \textbf{0.613} & \textbf{0.823} & 0.939 & \textbf{0.37} & \textbf{0.578} & \textbf{0.767} & 0.881\\
        \hline
        \multicolumn{1}{|c|}{2} & & 0.216 & 0.423 & 0.645 & 0.831 & 0.232 & 0.425 & 0.639 & 0.787\\
        \hline
        \multicolumn{1}{|c|}{2} & \cmark & 0.315 & 0.568 & 0.781 & 0.907 & 0.313 & 0.558 & 0.755 & 0.84\\
        \hline
    \end{tabular}}
    \caption{1-Recall@10 with different number of hidden layers (HL) and with 
    re-ranking with reconstructed features (RE) or not, while varying the number of bits in [64, 128, 256, 512].}
    \label{tab:res_asym_net_at10}
\end{table}

In this section, we analyze the effectiveness of the asymmetric search as well as the impact of difference choices of the hashing and decoder network structures. 
The asymmetric search has been described in Section~\ref{sec:asymmetric_search}. 
Here we use the candidate pool of 100 samples retrieved with the hamming distance, we compute the reconstruction of these samples from their hash codes and then use the Euclidean distance between the reconstructed features and the query feature to re-rank them.
We explore $3$ configurations of the hashing and decoder networks, namely no-hidden layers, 1 hidden layer and 2 hidden layers. 
Note that the hashing and decoder network have symmetric structures.
Specifically, the 1 hidden layer structure (1HL) has one fully connected layer of $8 \times n$ nodes and the 2 hidden layers structure (2HL) has one more fully connected layer of $8 \times m$ nodes, as detailed in Table~\ref{tab:architecture}.
The experiments conducted on the Deep1M dataset and the SIFT1M are reported in Table~\ref{tab:res_asym_net_at1} for the 1-Recall@1 and Table~\ref{tab:res_asym_net_at10} for the 1-Recall@10.

We can first observe that the network structure seem to have limited influence on the symmetric search (i.e. using only the hash codes), with all network configurations giving relatively similar performance for settings with the same number of bits. 
However, we observe that the 1HL network performs a bit better, especially with short hash codes (i.e. 64 and 128 bits).
Furthermore, we can see that the asymmetric search can significantly boost the retrieval performance on both datasets with the 1-Recall@1 approaching 50\% and the 1-Recall@10 of about 90\% when using $512$ bits.
The network with 2 hidden layers seem to have a lower performance than the model with only one hidden layer, that could be due to a slower convergence rate and thus the model not having fully converged in the allocated training time.

\subsection{Timing and compression analysis}

The average search time for one query within the database of 1 million samples using the HNSW structure is of $5$ms. 
The search time varies very little\footnote{The average query search time vary in a small range of $4-6$ms for all hash code lengths and methods tested.} 
when increasing the length of the hash codes, while each distance computation would cost slightly more.
This can be explained by the search time with HNSW being dominated by the exploration process (detailed in Section~\ref{sec:preliminary}) on the graph but not the distance computations.

Furthermore, longer hash codes usually induce a better traversal of the graph and thus reduce the number of distance computed in the search process. 
Note that the average number of distances computed for one query using the HNSW search procedure is in the range [10,000-20,000], while obviously the exhaustive search would compute 1 million distances.
The exhaustive search time, on the contrary, scales linearly with the hash code length, from about $7$ms for 64 bits hash codes to $31$ms for 512 bits.
The recall performance of the approximated search with HNSW is within $1\%$ of the exhaustive search recall.

\begin{table}[tb]
    \centering
    \resizebox{\columnwidth}{!}{\begin{tabular}{cc|c|c|c|c||c|c|c|c|}\cline{3-10}
        &  & \multicolumn{4}{c||}{Deep1M} & \multicolumn{4}{c|}{SIFT1M} \\
        \hline
        \multicolumn{1}{|c|}{HL} &  \multicolumn{1}{|c|}{GPU} & 64 & 128 & 256 & 512 & 64 & 128 & 256 & 512\\
        \hline
        \multicolumn{1}{|c|}{0} & & 0.888 & 1.006 & 1.062 & 1.194 & 1.002 & 1.061 & 1.096 & 1.240\\
        \hline
        \multicolumn{1}{|c|}{0} & \cmark & 1.283 & 1.291 & 1.358 & 1.447 & 1.287 & 1.374 & 1.409 & 1.490\\
        \hline
        \multicolumn{1}{|c|}{1} & & 2.184 & 2.284 & 2.519 & 2.922 & 2.348 & 2.533 & 2.759 & 3.361\\
        \hline
        \multicolumn{1}{|c|}{1} & \cmark & 1.504 & 1.453 & 1.484 & 1.640 & 1.426 & 1.450 & 1.578 & 1.581\\
        \hline
        \multicolumn{1}{|c|}{2} & & 3.657 & 4.913 & 7.585 & 14.862 & 4.047 & 5.612 & 8.785 & 16.939\\
        \hline
        \multicolumn{1}{|c|}{2} & \cmark & 1.733 & 1.639 & 1.710 & 1.875 & 1.725 & 1.743 & 1.779 & 1.947\\
        \hline
        
    \end{tabular}}
    \caption{Average reconstruction time (ms) of the top 100 samples retrieved using hashing models with different number of hidden layers (HL) and using a GPU (GPU) or not, while varying the number of bits in [64, 128, 256, 512].}
    \label{tab:rec_time}
\end{table}

We report the average time needed to reconstruct the top 100 samples found using the hash codes, with different model structures (no hidden layer, 1 or 2 hidden layers) and with or without GPU in 
Table~\ref{tab:rec_time}.
The CPU used when computing these timing was a 16 cores Intel(R) Xeon(R) E5630 @ 2.53GHz, while the GPU was a GeForce GTX TITAN X.
We can observe that the reconstruction time is evolving in a narrow range when using a GPU (about $1.3$ to $1.9$ms).
The timing on GPU seems slightly higher than the CPU one when using no hidden layers (likely due to the overhead of moving data to the GPU and back), but is much lower when using hidden layers.
Note that as our hashing model is shallow, even low-end GPU can be used to benefit from the speed-up of reconstruction.
After reconstruction, the re-ranking (of the top 100 reconstructed features) time is of about $0.1$ms.
As the encoding time of our method is of about $0.15$ms, overall our search process with reconstruction can be performed for any hash code size in less time than the exhaustive search while obtaining a much better recall.

\begin{table}[tb]
    \centering
    \resizebox{0.5\columnwidth}{!}{\begin{tabular}{|c|c|c|c|c|}
        \hline
        \# of bits & 64 & 128 & 256 & 512 \\
        \hline
        Deep1M & 48 & 24 & 12 & 6 \\
        \hline
        SIFT1M  & 64 & 32 & 16 & 8 \\
        \hline
        
    \end{tabular}}
    \caption{Compression rate on Deep1M and SIFT1M.}
    \label{tab:compression}
\end{table}

As Table~\ref{tab:compression} shows, our rank preserving hashing method achieves a significant compression ratio. 
Note that this compression ratio is the ratio between the storage of one original feature and its corresponding hash code. 
On Deep1M, original features are stored as 32-bit float numbers and on SIFT1M, original features are stored as 32-bit integers. 
A large-scale retrieval system based on HNSW graphs will benefit a lot from our hashing technique. 
For example, on SIFT1M, after the base set is compressed into 256-bits hash codes, we only need about $30.5$MB for all the nodes, which is more manageable (even on a low-end computing device) to be stored in RAM than the $488.3$MB of the original features. 
The total extra storage for the 1HL hashing model and decoder is only $3.2$MB, when the weights are stored as 32-bit float numbers.
Our hashing method enables to take the advantage of HNSW graphs for fast and accurate approximated nearest neighbour search on datasets of larger scale.

\section{Conclusions}

We have presented an unsupervised hashing method that aims to produce binary codes that preserve the ranking ability of an original real valued representation.
Our method clearly outperforms other hashing methods with publicly available implementations on two million-scale datasets Deep1M and SIFT1M, especially when hash code lengths are not too low.
Furthermore, we show how we can also learn a decoder enabling to reconstruct an approximation of original features and perform re-ranking without storing the original features.
This re-ranking step with reconstruction can significantly boost the retrieval performance.
Our proposed method thus enables compressing real-valued features into binary codes while preserving most of their retrieval ability.
We show that the proposed hashing and reconstruction method when used together with the popular hierarchical navigable small world graph indexing method can produce accurate search results at a very fast search speed.

\section*{Acknowledgements}
This research is based upon work supported by the 
Office of the Director of National Intelligence (ODNI), 
Intelligence Advanced Research Projects Activity (IARPA), 
via IARPA R\&D Contract No. 2014-14071600012.
The views and conclusions contained herein are those of  the authors and should not be interpreted as necessarily representing the official policies or endorsements, either expressed or implied, of the ODNI, IARPA, or the U.S. Government. 
The U.S. Government is authorized to reproduce and distribute reprints for Governmental purposes notwithstanding any copyright annotation thereon.

\begin{small}
\bibliographystyle{abbrv}
\bibliography{main-bib}

\begin{thebibliography}{10}

\bibitem{abadi2016tensorflow}
M.~Abadi, P.~Barham, J.~Chen, Z.~Chen, A.~Davis, J.~Dean, M.~Devin,
  S.~Ghemawat, G.~Irving, M.~Isard, et~al.
\newblock Tensorflow: a system for large-scale machine learning.
\newblock In {\em OSDI}, volume~16, pages 265--283, 2016.

\bibitem{babenko2016efficient}
A.~Babenko and V.~Lempitsky.
\newblock Efficient indexing of billion-scale datasets of deep descriptors.
\newblock In {\em Proceedings of the IEEE Conference on Computer Vision and
  Pattern Recognition}, pages 2055--2063, 2016.

\bibitem{bentley1975multidimensional}
J.~L. Bentley.
\newblock Multidimensional binary search trees used for associative searching.
\newblock {\em Communications of the ACM}, 18(9):509--517, 1975.

\bibitem{cai2016revisit}
D.~Cai.
\newblock A revisit of hashing algorithms for approximate nearest neighbor
  search.
\newblock {\em arXiv preprint arXiv:1612.07545}, 2016.

\bibitem{carreira2015hashing}
M.~A. Carreira-Perpin{\'a}n and R.~Raziperchikolaei.
\newblock Hashing with binary autoencoders.
\newblock In {\em Proceedings of the IEEE conference on computer vision and
  pattern recognition}, pages 557--566, 2015.

\bibitem{charikar2002similarity}
M.~S. Charikar.
\newblock Similarity estimation techniques from rounding algorithms.
\newblock In {\em Proceedings of the thiry-fourth annual ACM symposium on
  Theory of computing}, pages 380--388. ACM, 2002.

\bibitem{clevert2015fast}
D.-A. Clevert, T.~Unterthiner, and S.~Hochreiter.
\newblock Fast and accurate deep network learning by exponential linear units
  (elus).
\newblock {\em arXiv preprint arXiv:1511.07289}, 2015.

\bibitem{douze2016polysemous}
M.~Douze, H.~J{\'e}gou, and F.~Perronnin.
\newblock Polysemous codes.
\newblock In {\em European Conference on Computer Vision}, pages 785--801.
  Springer, 2016.

\bibitem{douze2018link}
M.~Douze, A.~Sablayrolles, and H.~J{\'e}gou.
\newblock Link and code: Fast indexing with graphs and compact regression
  codes.
\newblock In {\em Proceedings of the IEEE Conference on Computer Vision and
  Pattern Recognition}, pages 3646--3654, 2018.

\bibitem{duan2018minimizing}
L.-y. Duan, Y.~Wu, Y.~Huang, Z.~Wang, J.~Yuan, and W.~Gao.
\newblock Minimizing reconstruction bias hashing via joint projection learning
  and quantization.
\newblock {\em IEEE Transactions on Image Processing}, 27(6):3127--3141, 2018.

\bibitem{duan2018graphbit}
Y.~Duan, Z.~Wang, J.~Lu, X.~Lin, and J.~Zhou.
\newblock Graphbit: Bitwise interaction mining via deep reinforcement learning.
\newblock In {\em Proceedings of the IEEE Conference on Computer Vision and
  Pattern Recognition}, pages 8270--8279, 2018.

\bibitem{gong2013iterative}
Y.~Gong, S.~Lazebnik, A.~Gordo, and F.~Perronnin.
\newblock Iterative quantization: A procrustean approach to learning binary
  codes for large-scale image retrieval.
\newblock {\em IEEE Transactions on Pattern Analysis and Machine Intelligence},
  35(12):2916--2929, 2013.

\bibitem{he2018hashing}
K.~He, F.~Cakir, S.~A. Bargal, and S.~Sclaroff.
\newblock Hashing as tie-aware learning to rank.
\newblock In {\em 2018 IEEE/CVF Conference on Computer Vision and Pattern
  Recognition}, pages 4023--4032. IEEE, 2018.

\bibitem{heo2012spherical}
J.-P. Heo, Y.~Lee, J.~He, S.-F. Chang, and S.-E. Yoon.
\newblock Spherical hashing.
\newblock In {\em Computer Vision and Pattern Recognition (CVPR), 2012 IEEE
  Conference on}, pages 2957--2964. IEEE, 2012.

\bibitem{hu2018deep}
D.~Hu, F.~Nie, and X.~Li.
\newblock Deep binary reconstruction for cross-modal hashing.
\newblock {\em IEEE Transactions on Multimedia}, 2018.

\bibitem{hu2018hashing}
M.~Hu, Y.~Yang, F.~Shen, N.~Xie, and H.~T. Shen.
\newblock Hashing with angular reconstructive embeddings.
\newblock {\em IEEE Transactions on Image Processing}, 27(2):545--555, 2018.

\bibitem{ioffe2015batch}
S.~Ioffe and C.~Szegedy.
\newblock Batch normalization: Accelerating deep network training by reducing
  internal covariate shift.
\newblock {\em arXiv preprint arXiv:1502.03167}, 2015.

\bibitem{jegou2011product}
H.~Jegou, M.~Douze, and C.~Schmid.
\newblock Product quantization for nearest neighbor search.
\newblock {\em IEEE transactions on pattern analysis and machine intelligence},
  33(1):117--128, 2011.

\bibitem{jegou2012anti}
H.~J{\'e}gou, T.~Furon, and J.-J. Fuchs.
\newblock Anti-sparse coding for approximate nearest neighbor search.
\newblock In {\em Acoustics, Speech and Signal Processing (ICASSP), 2012 IEEE
  International Conference on}, pages 2029--2032. IEEE, 2012.

\bibitem{jegou2011searching}
H.~J{\'e}gou, R.~Tavenard, M.~Douze, and L.~Amsaleg.
\newblock Searching in one billion vectors: re-rank with source coding.
\newblock In {\em Acoustics, Speech and Signal Processing (ICASSP), 2011 IEEE
  International Conference on}, pages 861--864. IEEE, 2011.

\bibitem{kong2012isotropic}
W.~Kong and W.-J. Li.
\newblock Isotropic hashing.
\newblock In {\em Advances in neural information processing systems}, pages
  1646--1654, 2012.

\bibitem{krizhevsky2012imagenet}
A.~Krizhevsky, I.~Sutskever, and G.~E. Hinton.
\newblock Imagenet classification with deep convolutional neural networks.
\newblock In {\em Advances in neural information processing systems}, pages
  1097--1105, 2012.

\bibitem{kulis2009learning}
B.~Kulis and T.~Darrell.
\newblock Learning to hash with binary reconstructive embeddings.
\newblock In {\em Advances in neural information processing systems}, pages
  1042--1050, 2009.

\bibitem{kulis2009kernelized}
B.~Kulis and K.~Grauman.
\newblock Kernelized locality-sensitive hashing for scalable image search.
\newblock In {\em Computer Vision, 2009 IEEE 12th International Conference on},
  pages 2130--2137. IEEE, 2009.

\bibitem{li2018patternnet}
H.~Li, J.~G. Ellis, L.~Zhang, and S.-F. Chang.
\newblock Patternnet: Visual pattern mining with deep neural network.
\newblock In {\em Proceedings of the 2018 ACM on International Conference on
  Multimedia Retrieval}, pages 291--299. ACM, 2018.

\bibitem{liu2012supervised}
W.~Liu, J.~Wang, R.~Ji, Y.-G. Jiang, and S.-F. Chang.
\newblock Supervised hashing with kernels.
\newblock In {\em Computer Vision and Pattern Recognition (CVPR), 2012 IEEE
  Conference on}, pages 2074--2081. IEEE, 2012.

\bibitem{lowe2004distinctive}
D.~G. Lowe.
\newblock Distinctive image features from scale-invariant keypoints.
\newblock {\em International journal of computer vision}, 60(2):91--110, 2004.

\bibitem{malkov2018efficient}
Y.~A. Malkov and D.~A. Yashunin.
\newblock Efficient and robust approximate nearest neighbor search using
  hierarchical navigable small world graphs.
\newblock {\em IEEE transactions on pattern analysis and machine intelligence},
  2018.

\bibitem{rao2016diverse}
V.~Rao, P.~Jain, and C.~Jawahar.
\newblock Diverse yet efficient retrieval using locality sensitive hashing.
\newblock In {\em Proceedings of the 2016 ACM on International Conference on
  Multimedia Retrieval}, pages 189--196. ACM, 2016.

\bibitem{sablayrolles2017should}
A.~Sablayrolles, M.~Douze, N.~Usunier, and H.~J{\'e}gou.
\newblock How should we evaluate supervised hashing?
\newblock In {\em Acoustics, Speech and Signal Processing (ICASSP), 2017 IEEE
  International Conference on}, pages 1732--1736. IEEE, 2017.

\bibitem{salakhutdinov2009semantic}
R.~Salakhutdinov and G.~Hinton.
\newblock Semantic hashing.
\newblock {\em International Journal of Approximate Reasoning}, 50(7):969--978,
  2009.

\bibitem{song2015top}
D.~Song, W.~Liu, R.~Ji, D.~A. Meyer, and J.~R. Smith.
\newblock Top rank supervised binary coding for visual search.
\newblock In {\em Proceedings of the IEEE International Conference on Computer
  Vision}, pages 1922--1930, 2015.

\bibitem{song2015rank}
D.~Song, W.~Liu, D.~A. Meyer, D.~Tao, and R.~Ji.
\newblock Rank preserving hashing for rapid image search.
\newblock In {\em Data Compression Conference (DCC), 2015}, pages 353--362.
  IEEE, 2015.

\bibitem{wang2010sequential}
J.~Wang, S.~Kumar, and S.-F. Chang.
\newblock Sequential projection learning for hashing with compact codes.
\newblock In {\em Proceedings of the 27th international conference on machine
  learning (ICML-10)}, pages 1127--1134, 2010.

\bibitem{wang2013learning}
J.~Wang, W.~Liu, A.~X. Sun, and Y.-G. Jiang.
\newblock Learning hash codes with listwise supervision.
\newblock In {\em Proceedings of the IEEE International Conference on Computer
  Vision}, pages 3032--3039, 2013.

\bibitem{wang2013order}
J.~Wang, J.~Wang, N.~Yu, and S.~Li.
\newblock Order preserving hashing for approximate nearest neighbor search.
\newblock In {\em Proceedings of the 21st ACM international conference on
  Multimedia}, pages 133--142. ACM, 2013.

\bibitem{wang2018survey}
J.~Wang, T.~Zhang, N.~Sebe, H.~T. Shen, et~al.
\newblock A survey on learning to hash.
\newblock {\em IEEE Transactions on Pattern Analysis and Machine Intelligence},
  40(4):769--790, 2018.

\bibitem{wang2015ranking}
Q.~Wang, Z.~Zhang, and L.~Si.
\newblock Ranking preserving hashing for fast similarity search.
\newblock In {\em IJCAI}, pages 3911--3917, 2015.

\bibitem{weiss2009spectral}
Y.~Weiss, A.~Torralba, and R.~Fergus.
\newblock Spectral hashing.
\newblock In {\em Advances in neural information processing systems}, pages
  1753--1760, 2009.

\bibitem{yang2017discrete}
R.~Yang, Y.~Shi, and X.-S. Xu.
\newblock Discrete multi-view hashing for effective image retrieval.
\newblock In {\em Proceedings of the 2017 ACM on International Conference on
  Multimedia Retrieval}, pages 175--183. ACM, 2017.

\bibitem{zhao2015deep}
F.~Zhao, Y.~Huang, L.~Wang, and T.~Tan.
\newblock Deep semantic ranking based hashing for multi-label image retrieval.
\newblock In {\em Proceedings of the IEEE conference on computer vision and
  pattern recognition}, pages 1556--1564, 2015.

\end{thebibliography}
\end{small}

\end{document}